\documentclass[10pt,twocolumn,letterpaper]{article}

\newcommand{\eg}{\textit{e}.\textit{g}.}
\newcommand{\etal}{\textit{et al}.}
\newcommand{\ie}{\textit{i}.\textit{e}.}

\newcommand{\aka}{\textit{a.k.a.}}
\usepackage[table]{xcolor} 
\usepackage{times}
\usepackage{epsfig}
\usepackage{graphicx}
\usepackage{amsmath}
\usepackage{amssymb}
\usepackage{booktabs}
\usepackage{bbding}
\usepackage{multirow}
\usepackage{arydshln}

\usepackage{cvpr}              

\definecolor{cvprblue}{rgb}{0.21,0.49,0.74}
\usepackage[pagebackref,breaklinks,colorlinks,citecolor=cvprblue]{hyperref}

\def\paperID{10135} 
\def\confName{CVPR}
\def\confYear{2024}

\title{EventDance: Unsupervised Source-free Cross-modal Adaptation for Event-based Object Recognition }
\author{Xu Zheng$^{1}$ \quad Lin Wang$^{1}$$^{,2}$\thanks{Corresponding author.}\\
$^{1}$AI Thrust, HKUST(GZ)  \quad $^{2}$Dept. of CSE, HKUST \\
{\tt\small zhengxu128@gmail.com, linwang@ust.hk}
\\
\small{Project Page: \url{https://vlislab22.github.io/EventDance/}}
}

\begin{document}
\maketitle

\begin{abstract}

In this paper, we make the \textbf{first} attempt at achieving the cross-modal (\ie, image-to-events) adaptation for event-based object recognition \textbf{without accessing} any labeled source image data owning to privacy and commercial issues. Tackling this novel problem is non-trivial due to the novelty of event cameras and the distinct modality gap between images and events. In particular, as only the source model is available, a hurdle is how to extract the knowledge from the source model by only using the unlabeled target event data while achieving knowledge transfer. To this end, we propose a novel framework, dubbed \textbf{EventDance} for this unsupervised source-free cross-modal adaptation problem. Importantly, inspired by event-to-video reconstruction methods, we propose a reconstruction-based modality bridging (\textbf{RMB}) module, which reconstructs intensity frames from events in a self-supervised manner. This makes it possible to build up the surrogate images to extract the knowledge (\ie, labels) from the source model. We then propose a multi-representation knowledge adaptation (\textbf{MKA}) module that transfers the knowledge to target models learning events with multiple representation types for fully exploring the spatiotemporal information of events. The two modules connecting the source and target models are mutually updated so as to achieve the best performance. Experiments on three benchmark datasets with two adaption settings show that EventDance is on par with prior methods utilizing the source data.
\end{abstract}

\begin{figure}[t!]
    \centering
    \includegraphics[width=\linewidth]{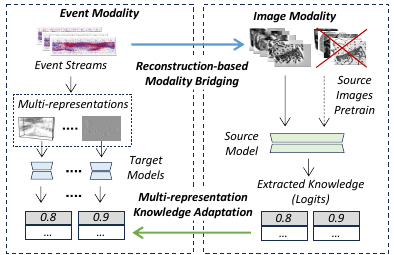}
    \caption{Illustration of the challenging task of the cross-modal adaptation from image to event modalities.
    We address it by introducing reconstruction-based modality bridging and multi-representation knowledge adaptation modules.}
    \label{fig:Teaser_figure}
\end{figure}

\section{Introduction}
Event cameras, \aka, the silicon retina~\cite{retinomorphic}, are bio-inspired novel sensors that perceive per-pixel intensity changes asynchronously and produce event streams encoding the time, pixel position, and polarity (sign) of the intensity changes~\cite{eventsurvey, EST, Hots}. Event cameras possess distinct merits, \eg, high dynamic range and no motion blur, making them more advantageous for challenging visual conditions, where the sensing quality degrades for the frame-based cameras. As a result, event cameras have recently drawn much attention from the computer vision and robotics community~\cite{deeplearningeventsurvey, RealtimeHSpeed, TORE, SITS, SpeedHDR, SLAMHSpeed, speedHDRrecon}. Although events are sparse and mostly encode the edge information, it has been shown that events alone are possible for learning scene understanding tasks, \eg, object recognition~\cite{eventsurvey}, via deep neural networks (DNNs). However, learning event-based DNNs is often impeded by the lack of large-scale precisely annotated datasets due to the asynchronous and sparse properties of event cameras, making it hardly possible to apply now-straightforward supervised learning.
For these reasons, some research endeavors have been made to explore the cross-modal adaptation to transfer knowledge from the labeled image modality (\ie, source) to the unlabeled event modality (\ie, target)~\cite{CTN,wang2021evdistill,messikommer2022bridging}.

In this paper, we make the \textbf{first} attempt to achieve cross-modal (\ie, image-to-event) adaptation for event-based object recognition where \textit{\textbf{we have no access to any labeled source image}}. The assumption of such a novel problem is of great importance for conditions where the labeled images, \ie, source modality data, \textit{are not allowed to be released} due to privacy and commercial issues~\cite{HTL, SHOT, MSFDA, ATheoretical}, and \textit{only the trained source models are shared}. However, tackling this problem is arduous due to 1) the sparse and asynchronous properties of events, making it difficult to directly apply existing cross-modal adaptation techniques,\eg,~\cite{SOCKET}, and 2) the significant modality gap between images and events as events mostly reflect edge information. In particular, as only the source modal is available, a hurdle is \textit{\textbf{how to extract the knowledge (\ie, pseudo label) from the source model by only using the unlabeled events and achieving knowledge transfer}}.

To this end, we formulate the learning objectives as \textbf{1)} bridging the modality gaps between image and event modalities and \textbf{2)} transferring the knowledge from the source image model to the target event domain. 
Accordingly, we propose a novel unsupervised source-free cross-modal adaptation framework, dubbed \textbf{EventDance}. Importantly, inspired by the attempt for event-to-video reconstruction~\cite{BackEventBasic}, we first propose a reconstruction-based modality bridging (\textbf{RMB}) module (Sec.~\ref{sec: Modality Bridging}), which builds up a \textbf{surrogate} image domain to imitate the source image distribution in a self-supervised manner. This allows for mitigating the large modality gap between images and events. 
Specifically, the RMB module takes an event stream to reconstruct multiple intensity frames to construct surrogate data in the image modality, which connects the source domain (inaccessible) of image modality. 
\textit{However, using~\cite{BackEventBasic} alone does not meet our needs as it only aims to reconstruct natural-looking images, not optimal surrogate images (inputs) for the source model}. Therefore, we optimize our RMB module for better knowledge extraction (\ie, pseudo labels) \textit{by minimizing the entropy of the source model's prediction} and ensuring \textit{temporal consistency} based on the high temporal resolution of event data.

Buttressed by the surrogate domain, we then propose a multi-representation knowledge adaptation (\textbf{MKA}) module (Sec.\ref{sec: Knowledge Adaptation}) that transfers the knowledge to learn target models for unlabeled events. As significant information loss, \eg, timestamp drops, occurs when converting events to a specific representation, like event count image, it may hinder the object recognition performance~\cite{deeplearningeventsurvey}. Therefore, we leverage several event representations in our EventDance, including stack image~\cite{stackimage}, voxel grid~\cite{voxelgrid}, and event spike tensor (EST)~\cite{EST}, to fully explore the spatio-temporal information of events. \textit{This allows for maintaining the cross-model prediction consistency training between the target models}. These two modules connecting the source and target models are mutually updated so as to achieve better modality bridging and knowledge adaptation.

\begin{figure}[t!]
    \centering
    \includegraphics[width=0.49\textwidth]{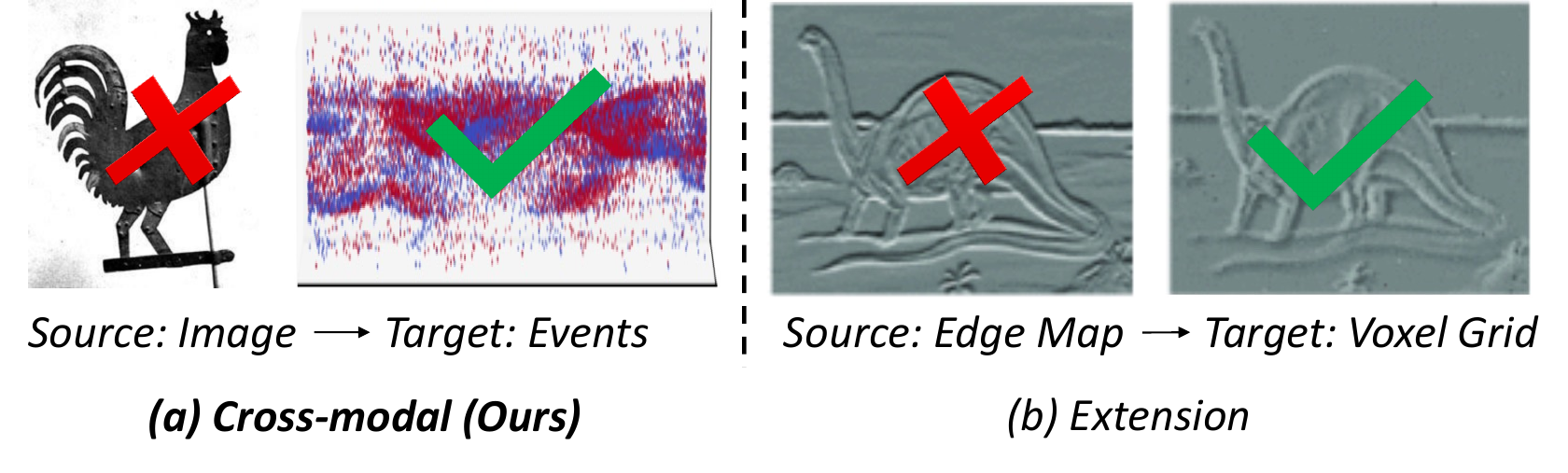}
    \caption{\textbf{Different adaptation settings}. 
    (a) ours from image to event modalities. (b) SFUDA from different image types~\cite{CTN}.}
    \label{fig:exp_set_com} 
\end{figure}

We validate EventDance on three event-based recognition benchmarks: N-Caltech 101~\cite{dataset}, N-MNIST~\cite{dataset}, and CIFAR10-DVS~\cite{li2017cifar10}. We show that EventDance performs well for the novel cross-modal (image-to-events) adaption task (see Fig.~\ref{fig:exp_set_com} \textcolor{red}{(a)}). We also show that it can be flexibly extended to a prior setting of \eg,~\cite{CTN}: edge map to event image adaptation (see Fig.\ref{fig:exp_set_com} \textcolor{red}{(b)}). The experimental results demonstrate that our EventDance significantly outperforms the prior source-free domain adaptation methods ~\eg, \cite{SHOT}, in addressing the challenging cross-modal task.
In summary, our main contributions are as follows: 
\textbf{(I)} We address a \textbf{novel yet challenging} problem for cross-modal (image-to-events) adaptation without access to the source image data.
\textbf{(II)} We propose EventDance, which incorporates the RMB and MKA modules to fully exploit event.
\textbf{(III)} Three event-based benchmarks with two adaptation settings 
 demonstrate the effectiveness and superiority of EventDance.

\section{Related Work}
\subsection{Event-based Object Recognition}
 aims to identify target objects from an event stream by taking full use of the event cameras' unique characteristics~\cite{zhou2024eventbind}. Since event cameras enjoy high temporal resolution, low latency, and very high dynamic range, this allows for real-time onboard object recognition in robotic, autonomous vehicles, and other mobile systems~\cite{deeplearningeventsurvey}.
 However, due to the distinct imaging paradigm shift, it is impossible to directly apply DNNs to learn events. Thus, various event representation types ~\cite{maqueda2018event, EvFlowNet, wang2019ev, deng2020amae, deng2021mvf, gehrig2019end, cannici2020differentiable, almatrafi2020distance, deng2022voxel, cao2023chasing} are proposed for mining the visual information and power from events, especially for the object recognition task. 
In previous works, \eg, ~\cite{CTN}, diverse event representations are adopted as the target domain while failing to explore the raw events. 
\textit{In this paper, we propose to learn target models that distinguish raw events and take multiple event representation for imposing consistency regularization. }

\begin{figure*}[t!]
    \centering
    \includegraphics[width=\textwidth]{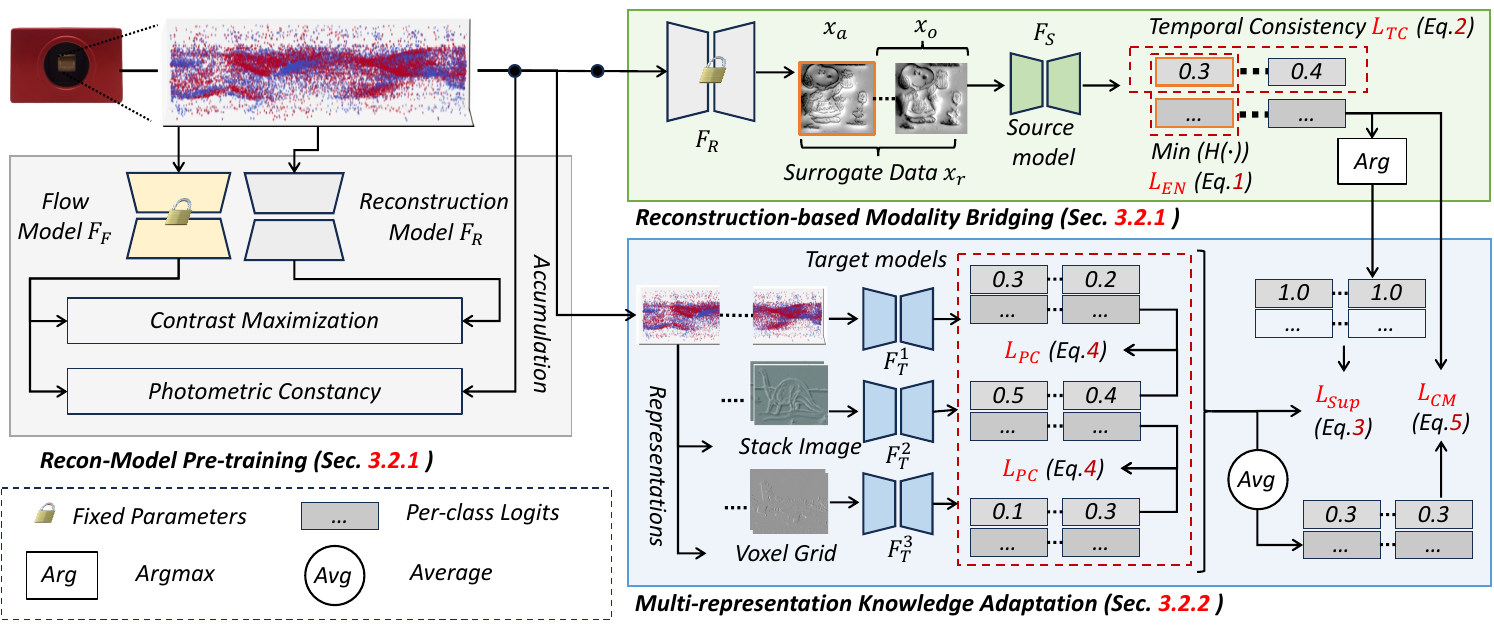}
    \vspace{-8pt}
    \caption{\textbf{Overall framework of our proposed framework}. RMB: reconstruction-based modality bridging module, MKA: multi-representation knowledge adaptation module. }
    \label{fig:overall}
\end{figure*}

\subsection{Cross-modal Knowledge Transfer}
Knowledge transfer across modalities is first proposed in ~\cite{CMKD}, aiming at learning representations for the modality with limited annotations based on a label-sufficient modality. Increasing attention has been paid to the cross-modal knowledge transfer task for novel sensors, \eg, event cameras. Most of the proposed methods~\cite{CMKDAR, CMKDAD, CMARD} assume that the cross-modal paired data is achievable while some recent works tried to relax this assumption by reducing the required data~\cite{KAP}. Additionally, there are some works for classification based on domain translation~\cite{DTRGBD, T2ARGBD}.
These methods rely on cross-modal data pairs and task-relevant paired data. To alleviate the demands on paired data, SOCKET~\cite{SOCKET} proposes an cross-modal adaptation framework that only utilizes external extra paired data for RGB-to-depth knowledge transfer, which are not always easy to obtain.
\textit{Differently, our EventDance is the \textbf{first} framework for cross-modal (image-to-event) adaptation without access to any source modality data, as shown in Fig.~\ref{fig:exp_set_com}~\textcolor{red}{(a)}. Due to the distinct modality gap with the image, we propose to build a surrogate domain and introducing representation consistency training for better knowledge transfer.}

\subsection{Source-free UDA}
UDA aims to alleviate the domain-shift problems caused by data distribution discrepancy in many computer vision tasks~\cite{zheng2023both, zheng2023look, CCUDA, LUDA, UDA-COPE, DUDA, DAWL, UMTDA, UDANAT, UDACAS, SUDA, zheng2024semantics}. However, the dependence on source data limits the generalization capability to some real applications, for reasons like data privacy issues~\cite{SHOT}. Thus endeavors have been made in transferring knowledge only from the trained source models~\cite{HTL} without access to the source data. The cross-domain knowledge for unlabeled target data is extracted from single~\cite{LPSUDA} or multiple~\cite{MSUDA} source models without access to the source data~\cite{SHOT}. The ideas of source-free UDA can be formulated into two types according to whether the parameters of source models are available~\cite{SFUDAsurvey}, \ie, white-box and black-box models. 
Concretely, the white-box are achieved by data generation~\cite{SFUDAsurrogate, SFUDAgenerate, SFDAestimation, UDAcontinual} and model fine-tuning~\cite {SFUDAdiversification, SFUDAbatchnorm} while the black-box depend on self-supervised learning~\cite{DINE, MutialNet} and distribution alignment~\cite{ITRL, DJP-MMD}. 

CTN~\cite{CTN} is a UDA framework that leverages the edge maps obtained from the source RGB images and adapts the classification knowledge to a target model learning event images, as shown in Fig.~\ref{fig:exp_set_com} \textcolor{red}{(b)}. 
In this paper, we focus on the source-free cross-modal (\ie, image-to-event) adaptation without accessing the source data, which is essentially different from~\cite{CTN} and more challenging to tackle. \textit{Our core idea is to create a surrogate domain in the image modality via the RMB module and update the surrogate domain for knowledge transfer via the MKA module.}

\section{The Proposed Framework}
\subsection{Problem Setup and Overview}
\label{sec:Problem Setup}

Knowledge adaptation from a source modality to a target modality can be more challenging than a domain shift between different datasets in the same modality, as demonstrated in~\cite{SOCKET}. 
Prior cross-modal adaption methods~\cite{DTRGBD, T2ARGBD, SOCKET} predominantly depend on extra data from both modalities to bridge the source and target modalities. 
However, the novelty of event cameras, combined with the lack of this paired data, impedes the application of these techniques to event modality.
Consequently, within the context of our cross-modal problem setup, we are limited to a pre-trained source model in the image modality and unlabeled target data in the event modality. 

\noindent \textbf{Our Key Idea}: \textit{By constructing the surrogate domain with target event data, we aim to mitigate modality gaps. This enables knowledge extraction from the source model. We subsequently employ multiple representations of the event data to accomplish the knowledge transfer.}

 \begin{figure}[t!]
    \centering
    \includegraphics[width=\linewidth]{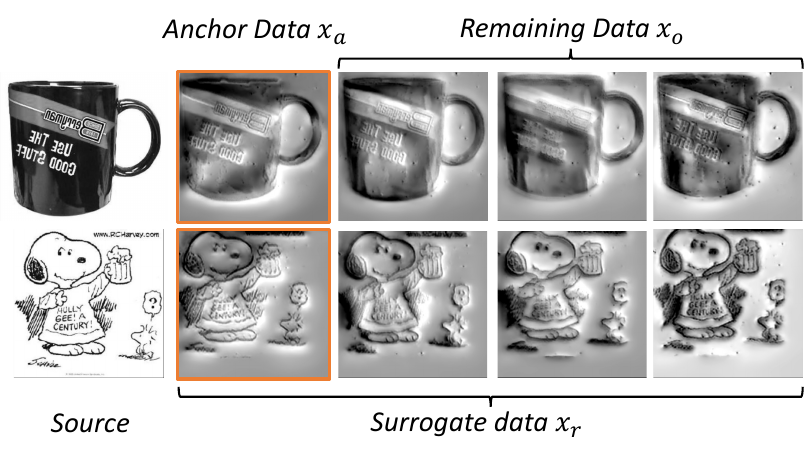}
    \caption{Visualization of samples in source and surrogate data in the image modality.}
    \label{fig:Recon_com}
\end{figure}

\noindent \textbf{Primary objective:} 
Denote the source model as $F_S$, where $S$ indicates the source modality on which the model is trained. $X_T$ represents the unlabeled target event data.
As shown in Fig.~\ref{fig:overall}, given a batch of event data $x_t \subset$ $X_T$, we can obtain the surrogate image batch $x_r$ using a reconstruction model $F_R$. 
Our aim is to derive target models $F_T^i$ using both $F_S$ and the unlabeled event data \textbf{$X_T$}. Here, the index $i$ indicates the $i$-th target model, which ingests different event representation forms as input. Specifically, for $i=1$, the input is a stack image; for $i=2$, it is a voxel grid; and for $i=3$, it is an event spike tensor (EST).
In the following sections, we elaborate the proposed modules: reconstruction-based modality bridging (RMB) module (Sec.~\ref{sec: Modality Bridging}) and multi-representation knowledge adaptation (MKA) module (Sec.~\ref{sec: Knowledge Adaptation}).
\subsection{Reconstruction-based Modality Bridging }
\label{sec: Modality Bridging}
The RMB module builds a surrogate image domain to imitate the source image distribution.
We utilize a self-supervised event-to-video model \cite{BackEventBasic} to construct the \textbf{surrogate data in the image modality} directly from raw event data, as depicted in Fig.~\ref{fig:Recon_com}. The surrogate data facilitates the extraction of knowledge (\ie, pseudo labels) from the image-trained source model.

\textit{However, simply introducing such a model is insufficient for our purpose}. The reason is that it only focuses on generating natural-looking images and not optimal surrogate images that can effectively extract knowledge from the source model.
Thus, we update the RMB module during training to generate better surrogate images for knowledge extraction. Specifically, we select the first surrogate image as the representative \textbf{anchor data} $x_a$, as shown in Fig.~\ref{fig:Recon_com}. Then, we pass it to the source model for predictions. Finally, we minimize the entropy of the prediction $F_{S}(x_a)$ to ensure that the surrogate images can effectively extract knowledge from the source model. We use $\mathcal{L}_{EN}$ to optimize the reconstruction model $F_R$,  which can be formulated as:
\begin{equation}
\label{l1}
\setlength{\abovedisplayskip}{3pt}
    \mathcal{L}_{EN} = min(H(F_S(x_a))), 
\setlength{\belowdisplayskip}{3pt}
\end{equation}
where $H(\cdot)$ represents the entropy function.  

 \begin{figure}[t!]
    \centering
    \includegraphics[width=\linewidth]{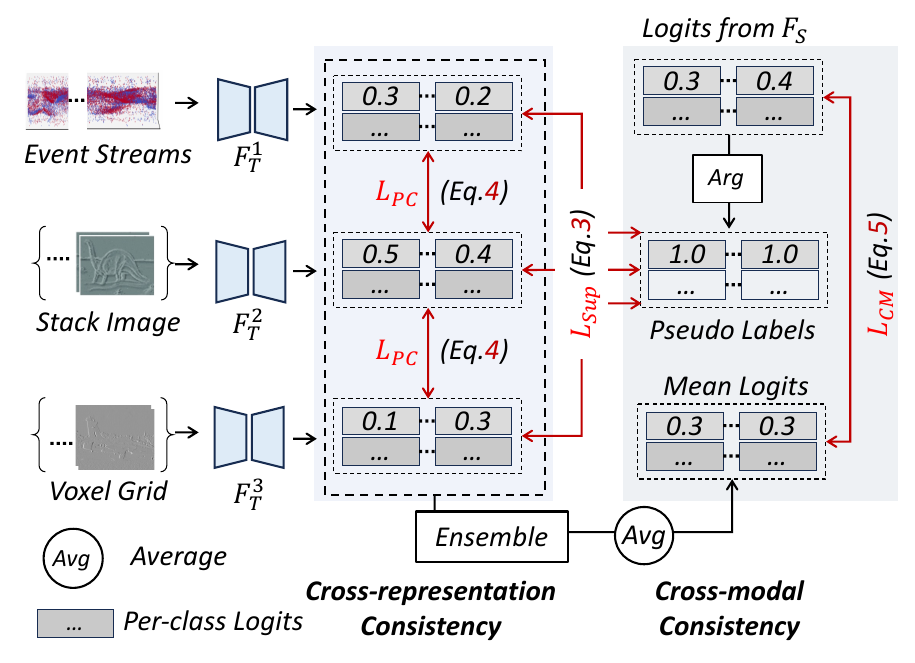}
    \caption{Illustration of the MKA module.}
    \label{fig:MB}
\end{figure}

As illustrated in Fig.~\ref{fig:Recon_com}, we also utilize the remaining data $x_o$ (excluding the selected anchor data $x_a$)  to augment the anchor data $x_a$ and fully leverage the high-temporal resolution of events. To ensure temporal prediction consistency among the reconstructed images, we update $F_{S}$ using the temporal consistency loss $\mathcal{L}_{TC}$:
\begin{equation}
\label{l2}
\setlength{\abovedisplayskip}{3pt}
    \mathcal{L}_{TC} = \mathcal{L}_{kl}(F_{S}(x_a),F_S(x_o)),
\setlength{\belowdisplayskip}{3pt}
\end{equation}
where $\mathcal{L}_{kl}$ is the Kullback-Liibler (KL) divergence.

In practice, we adopt the framework proposed in ~\cite{BackEventBasic} as our basic framework, which utilizes EvFlowNet~\cite{EvFlowNet} as the flow estimation model $F_F$ and E2VID~\cite{E2VID} as the reconstruction model $F_R$. EvFlowNet is trained with contrast maximization proxy loss~\cite{Contrast} and provides accurate optical flow estimation.  E2VID reconstructs intensity frames by exploring the flow-intensity relation with the event-based photometric constancy~\cite{Photometric}.
$F_F$ and $F_R$ are pre-trained using the unlabeled event data. Only $F_R$ is updated during training with $\mathcal{L}_{EN}$, while $F_F$ remains fixed to prevent $F_R$ from the model collapse.
With the knowledge extraction module, we obtain the prediction logits and pseudo labels $P$ from $F_S(x_a)$ for learning the target event-based models.

\subsection{Multi-representation Knowledge Adaptation}
\label{sec: Knowledge Adaptation}
Though our RMB module facilitates mitigating the modality gaps, adapting knowledge from images to events remains challenging due to: \textbf{1)} a single event representation type, \eg, voxel grid~\cite{CTN}, cannot comprehensively represent event data, leading to information loss during the adaptation process, and \textbf{2)} the source model is not ideal for cross-modal knowledge transfer, hindering the transfer efficiency. 
To this end, we propose to learn multiple target models using distinct event representations to fully leverage the high temporal resolution of events, as shown in Fig.~\ref{fig:overall}.

\begin{table*}[ht!]
\centering
\renewcommand{\tabcolsep}{14pt}
\resizebox{\linewidth}{!}{
\begin{tabular}{lcclcccl}
\toprule
Method & Backbone & N-MNIST & $\Delta$ & CIFAR10-DVS & $\Delta$ & N-CALTECH101 & $\Delta$ \\ \midrule
\multirow{3}{*}{Baseline} & R-18 & 41.80 & - & 36.14 & - & 40.81 & - \\ 
 & R-34 & 69.10 & - & 45.29 & - & 58.78 & - \\ 
& R-50 & 77.10 & - & 60.88 & - & 70.73 & - \\  \midrule
\multirow{3}{*}{SHOT ~\cite{SHOT}} & R-18 & 53.70 & +11.90 & 36.97 & +0.84  & 44.26 & +3.45 \\ 
 & R-34 &78.40 & +9.30 & 46.32 & +1.03 & 61.12 & +2.34 \\ 
&R-50 & 88.90 & +11.80 & 61.03 & +0.15 & 83.35 & +12.62 \\  \midrule
\multirow{3}{*}{Zhao \etal ~\cite{CTN}}& R-18 & 53.20 & +11.40 & 36.42 & +0.28 & 44.30 & +3.49 \\
 & R-34 & 76.90 & +7.80 & 45.78 & +0.49 & 61.10 & +2.32 \\ 
&  R-50 & 84.60 & +7.50 & 61.99 & +1.11 & 78.72 & +7.99 \\  \midrule
\multirow{3}{*}{SHOT++ ~\cite{SHOT++}}&R-18 & 68.80 & \underline{+27.00} & 37.26 & +1.12 & 49.33 & +8.52 \\ 
 & R-34 &84.70 & +15.60 & 46.37 & +1.08 & 64.54 & +5.76 \\ 
& R-50 & \underline{89.40} & +12.30 & 63.41 & +2.53 & \underline{82.88} & +12.15 \\  \midrule
& R-18 & 71.00 & \textbf{+29.20} & 62.13 & \underline{+25.99} & 66.77 & \textbf{+25.96}  \\ 
EventDance (Ours) & R-34 & 86.50 & +17.40 & \underline{71.98} & \textbf{+26.69} & 72.68 & +13.90 \\ 
& R-50 & \textbf{92.30} & +15.20 & \textbf{85.69} & +24.81 & \textbf{92.35} & \underline{+21.62} \\
 \bottomrule
\end{tabular}}
\caption{Experimental results on images-to-events with \textbf{SFUDA} methods (see Fig.~\ref{fig:exp_set_com}~\textcolor{red}{(a)}). 
$\Delta$: The performance gain over the baseline. 
The \textbf{bold} and \underline{underline} denote the best and the second-best performance in SFUDA methods, respectively.}
\label{tab: ourudasetting}
\end{table*}

To process raw event data, we convert a given event stream $E$ into commonly used event representations.
For voxel grid, as proposed in \cite{voxelgrid}, we obtain the voxel grid $E_v \subset \mathbb{R}^{H \times W \times C}$ with $B$ temporal bins using consecutive and non-overlapping segments of $E$, where $H$, $W$, and $C$ are spatial sizes. $E_v$ adaptively normalizes the temporal dimension of the input based on the timestamps of each segment of the event stream. For the event stack images, we employ a stacking strategy~\cite{stackimage} to sample and stack events in a fixed constant number. These results in a tensor-like representation $E_s \subset \mathbb{R}^{H \times W \times 1}$. For EST, we directly use the method in \cite{EST} taking raw events as input.

Technically, as shown in Fig.~\ref{fig:MB}, we set two training objectives for the target models as: \textbf{1)} \textit{cross-representation consistency} training with several event representation types in training target models $F_{T}^i$ and \textbf{2)} \textit{cross-modal consistency} training between the source model $F_{S}$ and $i$-th target model $F_{T}^i$. 
Based on the RMB module, the pseudo labels $P$ are obtained through the $argmax$ operation applied to $F_S(x_a)$.
We denote the $i$-th event representation for $F_{T}^i$ as $r(x_{t})^i$. 
The target models are supervised by the pseudo labels $P$ with the cross entropy $\mathcal{L}_{ce}$ as:
\begin{equation}
\label{l3}
\setlength{\abovedisplayskip}{3pt}  
    \mathcal{L}_{Sup} = \mathcal{L}_{ce}(F_{S}^i(r(x_{t})^i), P), i \in \{1,2,3\}.
\setlength{\belowdisplayskip}{3pt}   
\end{equation}
$\mathcal{L}_{Sup}$ serves as the fundamental knowledge transfer loss. 

\noindent \textbf{Cross-representation consistency.} These three event representation types are fed into their corresponding target models, and prediction consistency training is conducted among the models to facilitate target model learning from each other. The prediction consistency training loss among different event representation types can be formulated as:
\begin{equation}
\label{l4}
\setlength{\abovedisplayskip}{3pt}
    \mathcal{L}_{PC} = \sum^{3}_{k=1, l\neq k}\{\mathcal{L}_{kl}(F_T^{k}(r(x_{t})^{k}), F_T^{l}(r(x_{t})^{l}))\}.
\setlength{\belowdisplayskip}{3pt}
\end{equation} 

\noindent \textbf{Cross-modal consistency.} Additionally, as the source model is not an ideal model for image-to-event transfer, we propose a cross-modal consistency learning strategy between the source model $F_{S}$ and target model $F_{T}^i$ to simultaneously update both models. 
This improves the performance of $F_{S}$ and makes it more suitable for image-to-event knowledge transfer. The average ensemble results from $F_{T}^i$ and $F_{S}$ are mutually supervised by each other. The cross-modal consistency loss is formulated as follows:
\begin{equation}
\setlength{\abovedisplayskip}{3pt}
\setlength{\belowdisplayskip}{3pt}
\begin{aligned}
    \mathcal{L}_{\text{CM}} = &\mathcal{L}_{kl}(\frac{1}{3}\sum_{i=1}^3({F}_{T}^i(r(x_{t})^i)),F_{S}(x_a)) \\+ &\mathcal{L}_{kl}(F_{S}(x_a), \frac{1}{3}\sum_{i=1}^3({F}_{T}^i(r(x_{t})^i))).
\end{aligned}
\label{l5}
\end{equation}
Overall, our final loss is the combination of the above losses in Eq.~\ref{l1}, ~\ref{l2}, ~\ref{l3}, ~\ref{l4}, and ~\ref{l5}. The overall loss function $\mathcal{L}_{all}$: 
\begin{equation}
\setlength{\abovedisplayskip}{3pt}
\setlength{\belowdisplayskip}{3pt}
    \mathcal{L}_{all} = \mathcal{L}_{EN} + \mathcal{L}_{TC} + \mathcal{L}_{Sup} + \mathcal{L}_{PC} + \mathcal{L}_{CM}.
\end{equation}
Concretely, $\mathcal{L}_{EN}$ is used to optimize $F_R$; $\mathcal{L}_{TC}$ and $\mathcal{L}_{CM}$ are used to optimize $F_{S}$; and $\mathcal{L}_{Sup}$, $\mathcal{L}_{PC}$, and $\mathcal{L}_{CM}$ are used to optimize $F_{T}^i$. 
The whole framework is optimized in an end-to-end manner.

\section{Experiments}
In this section, we empirically validate various aspects of EventDance. In Sec.~\ref{sec: Datasets and Implementation Details}, we show the experimental settings of our image-to-event adaptation, baseline, comparison methods, and implementation details. We further show the performance of EventDance compared with the existing cross-modal and UDA methods in Sec.~\ref{sec: Experimental Results}.
\subsection{Datasets and Implementation Details}
\label{sec: Datasets and Implementation Details}
\noindent \textbf{N-MNIST}\cite{dataset} is the event-based version of the well-known MNIST dataset. The dataset was created by capturing the visual input of an event camera focused on a monitor displaying the original MNIST data. \textbf{N-CALTECH101}\cite{dataset} is the event-based extension of the CALTECH101 dataset, which contains 100 object classes along with a background class. This dataset poses a challenge due to the many classes and the unbalanced number of samples within each class. \textbf{CIFAR10-DVS}~\cite{li2017cifar10} consists of 10,000 event streams belonging to 10 classes, captured by a DVS camera with a spatial resolution of 128 × 128.\\
\begin{table}[t!]
\centering
\renewcommand{\tabcolsep}{6pt}
\resizebox{\linewidth}{!}{
\begin{tabular}{lccccc}
\toprule 
Method & S.F. & Unsup. & Backbone / Train & N-CAL \\ \midrule
E2VID~\cite{E2VID} & \XSolidBrush & \Checkmark & Fine-tune & 59.80 \\ 
 + CLIP& \XSolidBrush & \Checkmark & Scratch & 9.40 \\ \midrule
Ev-LaFOR~\cite{cho2023label} & \XSolidBrush & \Checkmark  & Text Prompt & 82.46  \\ 
 + CLIP& \XSolidBrush & \Checkmark & Visual Prompt & 82.61 \\ 
\midrule
\multirow{3}{*}{Wang \etal~\cite{wang2019event}} & \Checkmark & \Checkmark & - & 42.70 \\
& \XSolidBrush & \Checkmark & - & 43.50  \\ 
 + CLIP & \Checkmark & \XSolidBrush & - & 39.70 \\ \midrule
\multirow{3}{*}{DSAN~\cite{DSAN}}& \XSolidBrush & \Checkmark& R-18 & 78.45 \\
& \XSolidBrush & \Checkmark & R-34 & 89.01  \\ 
& \XSolidBrush & \Checkmark & R-50 & 94.56 \\ \midrule
& \Checkmark & \Checkmark & R-18 & 66.77  \\ 
EventDance (Ours) & \Checkmark & \Checkmark & R-34 & 72.68 \\ 
& \Checkmark & \Checkmark & R-50 & 92.35 \\ 
\bottomrule
\end{tabular}}
\caption{Experimental results compared with label-free methods.}
\label{tab:labelfree}
\end{table}
\begin{table}[t!]
\centering
\renewcommand{\tabcolsep}{8pt}
\resizebox{\linewidth}{!}{
\begin{tabular}{lcccc}
\toprule 
Method & S.F. & Unsup. & Backbone & N-MNIST \\ \midrule
EV-VGCNN~\cite{deng2022voxel} & \XSolidBrush & \XSolidBrush & EV-VGCNN & 99.10   \\ 
Deep SNN~\cite{lee2016training} & \XSolidBrush & \XSolidBrush & Deep SNN & 98.70 \\
Phased LSTM~\cite{neil2016phased} & \XSolidBrush & \XSolidBrush & Phased LSTM & 97.30 \\
PointNet++~\cite{wang2019space} & \XSolidBrush & \XSolidBrush & PointNet++ & 95.50 \\
\midrule
& \Checkmark & \Checkmark & R-18 & 71.00  \\ 
EventDance (Ours) & \Checkmark & \Checkmark & R-34 & 86.50 \\ 
& \Checkmark & \Checkmark & R-50 & 92.30 \\ 
\bottomrule
\end{tabular}}
\caption{Experimental results compared with supervised methods.}
\label{tab:supervised}
\end{table}
\noindent \textbf{Evaluation configurations.}
EventDance employs three target models by taking three event representation types (stack image, voxel grid, and EST) as inputs in the training phase, respectively. Therefore, \textit{the inference can be achieved using one of the models}. 
We present the recognition accuracy results of the target models taking voxel gird, on three event-based benchmarks in Tab.~\ref{tab: ourudasetting}. 

\noindent \textbf{Baseline and comparison methods.}
As we are the first to address the cross-modal problem, there is no direct baseline available for comparison. We establish a baseline in Tab.~\ref{Tab: Edge2Vox} and Tab.~\ref{tab: ourudasetting} to evaluate the performance of the pre-trained source model with event voxel grids, as in the previous work~\cite{CTN}. Also, we compare our method with the existing SFUDA methods~\cite{SHOT, SHOT++}, image-to-voxel-grid adaptation method~\cite{CTN}, and the UDA method using source data~\cite{DSAN} that use event voxel grids as the target modality data. 
\\
\noindent \textbf{Implementation Details.}
We use ResNet-18, 34, and 50 (R-18, R-34, and R-50) pre-trained on ImageNet as backbones. The batch size is set to 64, following the prior work~\cite{CTN}. We use the AdamW optimizer with a learning rate of 1e-5, which linearly decays over time. We use image augmentation techniques, \eg, random rotations and flipping for source modality pre-training. However, we do not use event augmentation techniques during target learning for a fair comparison with other methods. 
\textit{More details about the settings can be found in the supplmat.}
\begin{figure}[t!]
    \centering
    \includegraphics[width=\linewidth]{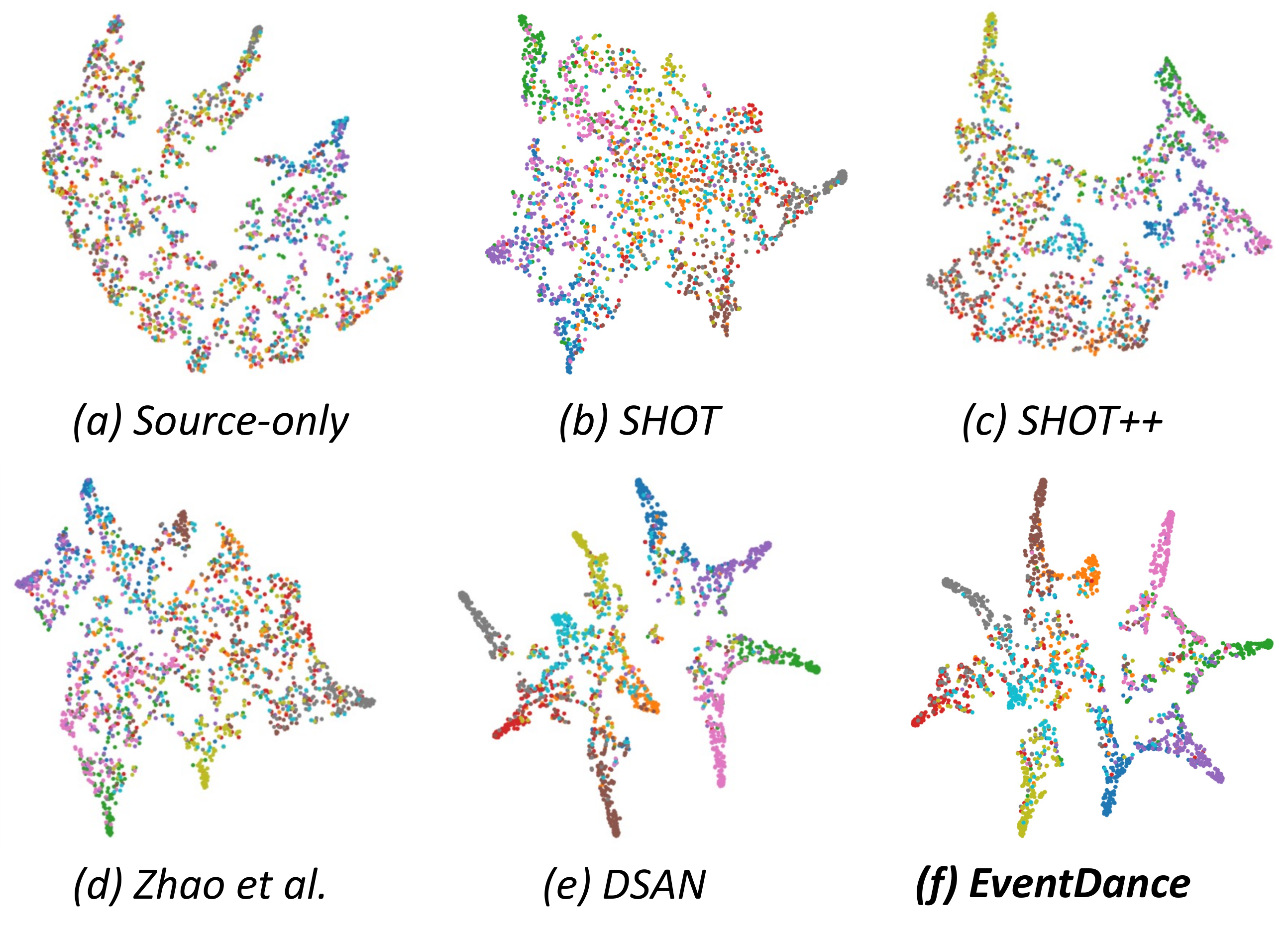}
    \caption{TSNE~\cite{van2008visualizing} visualization of (a) source-only, (b) SHOT, (c) SHOT++, (d) Zhao \etal, (e)DSAN, and (f) EventDance on the target modality CIFAR10-DVS dataset with R-18 backbone. Different colors represent the 10 classes in CIFAR10-DVS dataset.}
    \label{fig:tsne}
\end{figure}

\subsection{Experimental Results}
\label{sec: Experimental Results}
We evaluate our EventDance under the challenging source-free image-to-events adaptation setting. 
The experimental results are shown in Tab.~\ref{tab: ourudasetting}. 
EventDance consistently outperforms the source-free UDA methods~\cite{SHOT,SHOT++}, source-free cross-modal UDA method~\cite{CTN} and even achieves recognition accuracy closer to that of the UDA method DSAN~\cite{DSAN} that utilizes the source data on the three event-based benchmarks. 
EventDance brings significant performance gains of +25.99\%, +26.69\%, and +24.81\% with R-18, R-34, and R-50 backbones, respectively. This indicates the superiority of our proposed RMB and MKA modules in tackling the non-trivial cross-modal problem.

As shown in Tab.~\ref{tab:labelfree}, we also compare our EventDance with the label-free methods, such as E2VID~\cite{E2VID} + CLIP, Ev-LaFOR~\cite{cho2023label} + CLIP, Wang \etal~\cite{wang2019event} + CLIP, and UDA method DSAN~\cite{DSAN}. Obviously, our EventDance outperforms these label-free methods and achieves recognition accuracy that is closer to that of the UDA method DSAN (with source data)~\cite{DSAN}, even without using source modality data (92.35\% vs. 94.56\% with R-50 backbone).

Furthermore, we provide a performance comparison between our EventDance and several state-of-the-art supervised event-based recognition methods on the N-MNIST dataset in Tab.~\ref{tab:supervised}, including EV-VGCNN~\cite{deng2022voxel}, Deep SNN~\cite{lee2016training}, Phased LSTM~\cite{neil2016phased}, and PointNet++~\cite{wang2019space}. Our EventDance achieves good performance in an unsupervised manner, without using the source data.
We provide the TSNE~\cite{van2008visualizing} visualization in Fig.~\ref{fig:tsne}, apparently, our EventDance brings a significant improvement in distinguishing cross-modal samples in high-level feature space. 

\section{Ablation Study and Analysis}
\label{sec: Ablation Study and Analysis}
\begin{figure}[t!]
    \centering
    \includegraphics[width=\linewidth]{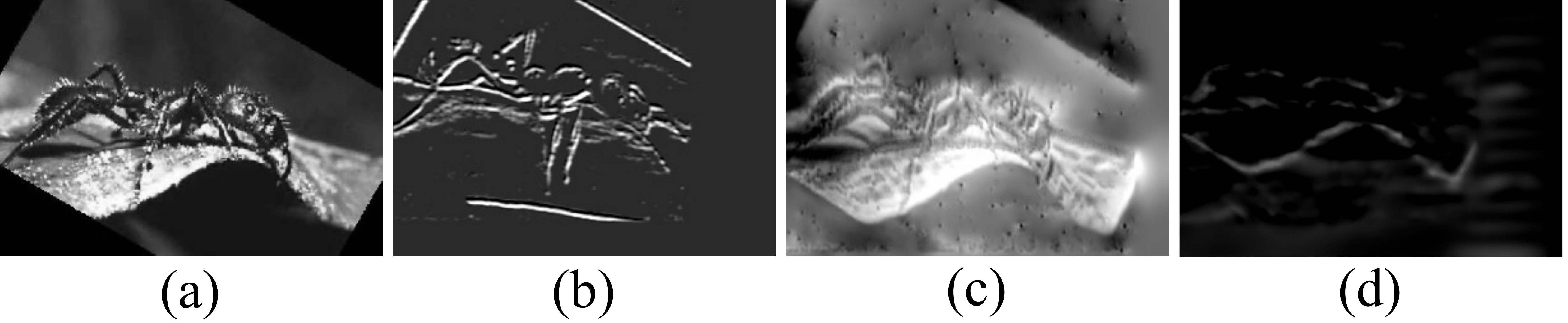}
    \caption{Visualization of (a) source gray-scale image; (b) event stack image; (c) reconstructed image before training; (d) reconstructed image after training.}
    \label{fig:Recon_training}
\end{figure}

\noindent \textbf{Different combination of proposed modules.}
To validate the effectiveness of the proposed modules, we conduct experiments on the NCALTECH-101 dataset with different combinations of modules. Tab.~\ref{tab: modulecombin} shows the detailed results of the performance with different loss and component combinations. All of our proposed modules and loss functions have a positive impact on improving recognition accuracy. Notably, fine-tuning the reconstruction model $F_R$ results in a significant performance gain by 17.83\%, which supports our claim of building a surrogate data in the image modality for better knowledge transfer, rather than the visual quality of event-to-video reconstruction. This is further supported by the visual results presented in Fig.~\ref{fig:Recon_training}. 

\begin{table}[t!]
\centering
\setlength{\tabcolsep}{6pt}
\resizebox{\linewidth}{!}{
\begin{tabular}{cccccccc}
\toprule
$F_R$&$\mathcal{L}_{TC}$&$\mathcal{L}_{EN}$&$\mathcal{L}_{Sup}$&$\mathcal{L}_{PC}$&$\mathcal{L}_{CM}$&Accurracy&$\Delta$
 \\ \midrule
 &  &  & \Checkmark &  &  & 40.81 & - \\ 
 &  & \Checkmark & \Checkmark &  &  & 43.36 & +2.55 \\ 
 & \Checkmark &  & \Checkmark &  &  & 45.89 & +5.08 \\ 
\Checkmark &  &  & \Checkmark &  &  & 58.64 & +17.83 \\ 
 & \Checkmark & \Checkmark & \Checkmark &  &  & 53.40 & +10.04 \\ 
\Checkmark & \Checkmark & \Checkmark & \Checkmark &  &  & 61.83 & +21.02 \\
 &  &  & \Checkmark &  & \Checkmark  & 43.38 & +2.57 \\ 
  &  &  & \Checkmark & \Checkmark &   & 42.56 & +1.75 \\ 
\Checkmark & \Checkmark & \Checkmark & \Checkmark &  & \Checkmark  & 63.58 & +22.77 \\ 
\Checkmark & \Checkmark & \Checkmark & \Checkmark & \Checkmark &  & 63.26 & +22.45 \\ 
\Checkmark & \Checkmark & \Checkmark & \Checkmark & \Checkmark & \Checkmark &66.77 & +25.96  \\ \bottomrule
\end{tabular}}
\caption{Ablation study of different module combinations on N-CALTECH101 with ResNet-18.}
\label{tab: modulecombin}
\end{table}

\begin{table}[t!]
\centering
\setlength{\tabcolsep}{10pt}
\resizebox{\linewidth}{!}{
\begin{tabular}{cccc}
\toprule
\multirow{2}{*}{Backbone} & \multicolumn{3}{c}{Event Representations} \\ \cmidrule{2-4} 
 & Stack Image & Voxel Grid & Event Spike Tensor \\ \midrule
R-18 & $66.70_{\textcolor{blue}{-0.07}}$ & 66.77 & $66.96_{\textcolor{red}{+0.19}}$ \\ \midrule
R-34 & $71.16_{\textcolor{blue}{-1.52}}$ & 72.68 & $73.00_{\textcolor{red}{+0.52}}$ \\ \midrule
R-50 & $91.54_{\textcolor{blue}{-0.81}}$ & 92.35 & $92.74_{\textcolor{red}{+0.39}}$  \\ \bottomrule
\end{tabular}}
\caption{Ablation experiments on the inference of our proposed method with different event representations.}
\label{Tab: TargetRep}
\end{table}
\begin{table*}[t!]
\centering
\setlength{\tabcolsep}{6pt}
\resizebox{\textwidth}{!}{
\begin{tabular}{cccccccccc}
\toprule
Method & Source-Free & Unsupervised & Backbone & N-MNIST & $\Delta$ & CIFAR10-DVS & $\Delta$ & N-CALTECH101 & $\Delta$ \\ \midrule
\multirow{3}{*}{Baseline} && & R-18 & 82.60 & - & 54.10 & - & 72.80 & - \\ 
 & \XSolidBrush & \Checkmark & R-34 & 84.30 & - & 55.70 & - & 73.20 & - \\ 
 &&& R-50 & 84.70 & - & 56.50 & - & 74.70 & - \\  
 \midrule
\multirow{3}{*}{Zhao \etal ~\cite{CTN}} &&& R-18 & 98.60 & +16.00 & 76.50 & +22.40 & 88.50 & +15.70 \\ 
 & \Checkmark & \Checkmark  & R-34 & 99.00 & +14.70 & 76.70 & +21.00 & 89.30 & +16.10 \\ 
&& & R-50 & 99.30 & +14.60 & 77.30 & +20.80 & 90.10 & +15.40 \\  
 \midrule
&& & R-18 & 99.10 & \textbf{+16.50} & 79.80 & +25.70  & 90.30 & +17.50 \\  
EventDance (ours) & \Checkmark & \Checkmark & R-34 & \underline{99.40} & \underline{+15.10}  & \underline{85.40} & \underline{+29.70}  & \underline{91.40} & \textbf{+18.20}  \\
& & & R-50 & \textbf{99.70} & +15.00  & \textbf{86.70} & \textbf{+30.20}  & \textbf{92.30} & \underline{+17.60}  \\ 
 \midrule
\end{tabular}}
\caption{Experimental results on edge maps-to-voxel grids with UDA methods (see Fig.~\ref{fig:exp_set_com} \textcolor{red}{(a)}). Test Rep.: event representation used in test; VG: voxel grid. The \textbf{bold} and \underline{underline} denote the best and the second-best performance in source-free uda methods, respectively.}
\label{Tab: Edge2Vox}
\end{table*}

\noindent\textbf{Event representation vs. target model's performance.}
For a fair comparison, we validate the quantitative results of all methods in Tab.~\ref{Tab: Edge2Vox} and Tab.~\ref{tab: ourudasetting} using event voxel grids. To investigate how to fully leverage the abundant spatio-temporal information of events for object recognition, we provide the results of validating our EventDance with different representation types in Tab.~\ref{Tab: TargetRep}. 
Compared to inference with voxel grids, using EST, which contains more spatio-temporal information of events, achieves the best recognition accuracy gains by +0.19\%, +0.52\%, and +0.39\%, on the backbones R-18, R-34, and R-50, respectively. The results reflect that EST is better suited for object recognition.
The stack images, for which temporal information is lost, achieve lower recognition accuracy than the voxel grids by -0.07\%, -1.52\%, and -0.81\% with backbones R-18, R-34, and R-50, respectively. Thus, it is crucial to explore the event data with representations that remain temporal characteristic of events in the cross-modal adaptation problems.

\begin{table}[]
\centering
\setlength{\tabcolsep}{13pt}
\resizebox{\linewidth}{!}{
\begin{tabular}{ccccc}
\toprule
\begin{tabular}[c]{@{}c@{}}Representation\end{tabular} & S & V & E & All \\ \midrule
Accuracy & 61.63 & 63.58 & 65.74 & \textbf{66.91} \\ \bottomrule
\end{tabular}}
\caption{Ablation on the usage of event representations in target model training. (S: stack image; V: voxel grid; E: EST.)}
\label{Tab: repcomnbin}
\end{table}



\noindent \textbf{Ablation of RMB module.}
The RMB module is a crucial component for bridging the modality gaps between images and events.
The results are shown in Tab.~\ref{tab: modulecombin}, where we use a checkmark to denote $F_R$ fine-tuning. As can be seen, only using the pre-trained reconstruction model $F_R$ to construct the surrogate data achieves the recognition accuracy of 53.04\%. Obviously, updating $F_R$ improves the accuracy to 61.83\%. We visualize the surrogate data's intensity frames, as shown in Fig.~\ref{fig:Recon_training}. Although the reconstructed image in Fig.~\ref{fig:Recon_training}~\textcolor{red}{(d)} after training is less distinct than the one in Fig.~\ref{fig:Recon_training}~\textcolor{red}{(c)} before training, the ability to extract knowledge becomes significantly better (53.04\% vs. 61.83\%).

\section{Extension Experiment}
We show that our method can be flexibly extended to the adaptation problem from edge maps to event voxel grids, as done in~\cite{CTN}. Specifically, we test our EventDance under the source-free UDA setting~\cite{CTN}, where event streams are converted into event voxel grids with $C=3$ as the target data, and images are processed to edge maps as source modality data for pre-training source models. 
As in Tab.~\ref{Tab: Edge2Vox}, our approach consistently outperforms the SoTA method CTN~\cite{CTN} on three event-based benchmarks. 
\section{Discussion}
\label{Discussion}

\noindent \textbf{Performance gains in two experimental settings.} Edge maps have a closer similarity to voxel grids of events, which makes the knowledge transfer easier compared to our cross-modal setting. Therefore, our EventDance achieves a greater performance gain in the image-to-events setting (26.15\% w/ R-18) compared to the edge maps-to-voxel grids (+17.50\% w/ R-18) on NCALTECH-101. 
 
\noindent \textbf{High temporal resolution of events.}
Cross-modal knowledge transfer is challenging due to the distinct modality gap between images and events, namely $H \!\times\! W \!\times\! C$ for images and $(x,y,t,p)$ for events~\cite{deeplearningeventsurvey}. The straightforward approach to alleviating this problem is to convert events into image-like tensors. However, most event representations struggle with information loss, such as temporal information. For downstream tasks, there might be the best event representation that achieves the SoTA performance, such as EST~\cite{EST} in object recognition. Nevertheless, we find that multiple event representations are suitable for source-free cross-modal adaptation. This observation is supported by the quantitative results shown in Tab.~\ref{Tab: repcomnbin}.

\noindent \textbf{Surrogate data in training.}
While building surrogate data incurs higher computation costs, it effectively eliminates the need for extra paired data that may not always be available in practice. Moreover, in our work, the reconstruction model used to construct the surrogate data is only trained and updated during the training phase and can be freely discarded during the inference.

\noindent \textbf{Selection of anchor data.}
We experimentally determine to select the first surrogate image as anchor data, which is reconstructed from the initial period of the event stream.
This is the most effective method to obtain reliable anchor data, as the size of event streams varies across the target dataset. 
The remaining frames are treated as augmentation for anchor data. Attempts at random selection result in low-quality images for shorter streams.

\section{Conclusion}
In this paper, we investigated a new problem of achieving image-to-event adaptation for event-based object recognition without access to any source image. 
To this end, we proposed an cross-modal framework, named EventDance. 
The experiments for image-to-events and edge maps-to-voxel grids adaptation show that EventDance outperforms prior source-free and cross-modal UDA methods and is on par with the methods that use source data. 

\noindent \textbf{Limitation and Future Work:} 
One limitation of EventDance is that training three target models with different representations lead to increased computational costs during training. However, our method has significant implications for the event-based vision and may open a new research direction. In the future, we plan to extend our approach to other downstream tasks. 

\noindent \textbf{Acknowledgement}
This paper is supported by the National Natural Science Foundation of China (NSF) under Grant No. NSFC22FYT45 and the Guangzhou City, University and Enterprise Joint Fund under Grant No.SL2022A03J01278.
\clearpage
{
    \small
    \bibliographystyle{ieeenat_fullname}
    \bibliography{main}
}


\end{document}